\documentclass[10pt,twocolumn,letterpaper]{article}

\usepackage[pagenumbers]{cvpr} 

\usepackage{graphicx}
\usepackage{amsmath}
\usepackage{amssymb}
\usepackage{booktabs}
\usepackage[accsupp]{axessibility}
\usepackage[pagebackref,breaklinks,colorlinks]{hyperref}

\usepackage[capitalize]{cleveref}
\crefname{section}{Sec.}{Secs.}
\Crefname{section}{Section}{Sections}
\Crefname{table}{Table}{Tables}
\crefname{table}{Tab.}{Tabs.}

\def\cvprPaperID{5973} 
\def\confName{CVPR}
\def\confYear{2023}

\begin{document}

\title{Exploring Discontinuity for Video Frame Interpolation}

\author{Sangjin Lee$^{*,1}$ \quad Hyeongmin Lee$^{*,1}$ \quad Chajin Shin$^1$ \quad Hanbin Son$^1$ \quad Sangyoun Lee$^{1,2}$\\
$^1$~Yonsei University\\
$^2$~Korea Institute of Science and Technology (KIST)\\
\{sglee97, minimonia, chajin, hbson, syleee\}@yonsei.ac.kr
}

\maketitle
\def\thefootnote{*}\footnotetext{Both authors contributed equally to this work.}

\begin{abstract}
Video frame interpolation~(VFI) is the task that synthesizes the intermediate frame given two consecutive frames. Most of the previous studies have focused on appropriate frame warping operations and refinement modules for the warped frames. These studies have been conducted on natural videos containing only continuous motions. However, many practical videos contain various unnatural objects with discontinuous motions such as logos, user interfaces and subtitles. We propose three techniques to make the existing deep learning-based VFI architectures robust to these elements. First is a novel data augmentation strategy called figure-text mixing~(FTM) which can make the models learn discontinuous motions during training stage without any extra dataset. Second, we propose a simple but effective module that predicts a map called discontinuity map~($D$-map), which densely distinguishes between areas of continuous and discontinuous motions. Lastly, we propose loss functions to give supervisions of the discontinuous motion areas which can be applied along with FTM and $D$-map. We additionally collect a special test benchmark called Graphical Discontinuous Motion~(GDM) dataset consisting of some mobile games and chatting videos. Applied to the various state-of-the-art VFI networks, our method significantly improves the interpolation qualities on the videos from not only GDM dataset, but also the existing benchmarks containing only continuous motions such as Vimeo90K, UCF101, and DAVIS.
\end{abstract}

\section{Introduction}
Video frame interpolation~(VFI) task is to generate the intermediate frame given some consecutive frames from a video. When the time interval of each input frames is fixed, we can get smoother video, and when the frame rate is fixed, we can get slow-motion video. This can also be applied to other vision tasks such as video compression~\cite{bao2018high,wu2018video}, view synthesis~\cite{flynn2016deepstereo,zhou2016view,kalantari2016learning}, and other real-world applications~\cite{paikin2021efi,Yu2021ICCV,siyao2021deep}.

\begin{figure}[t]
	\setlength{\belowcaptionskip}{-15pt}
	\begin{center}
		\includegraphics[width=0.99\linewidth]{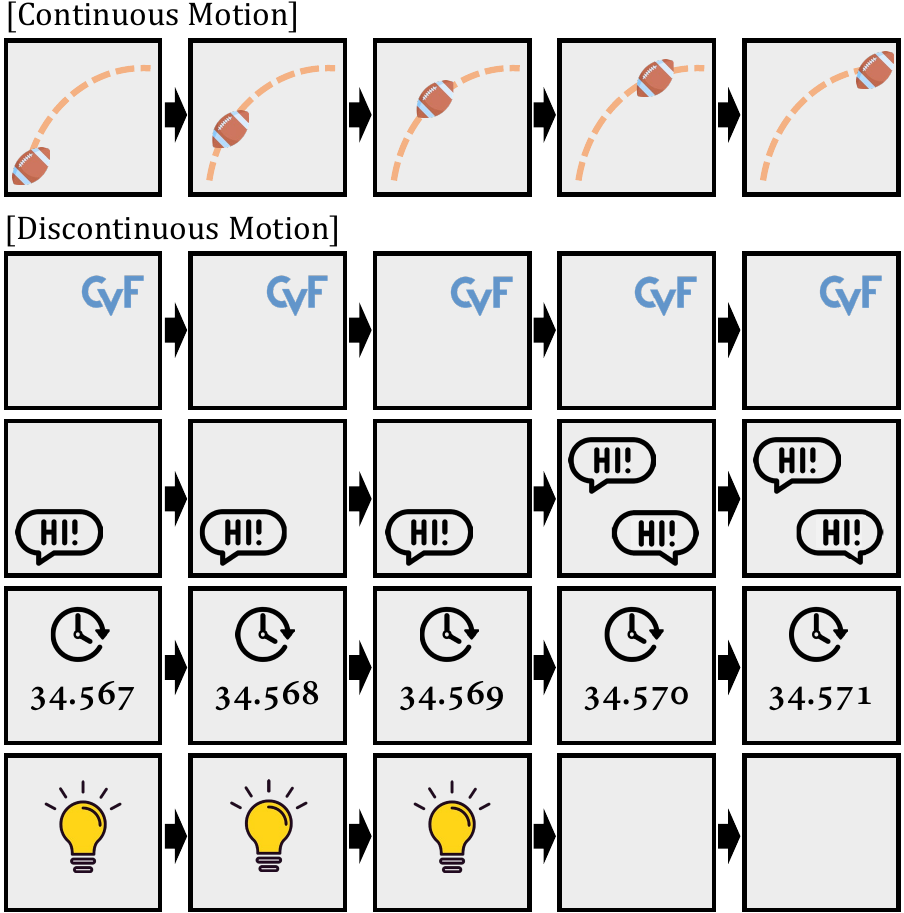}
	\end{center}
	\vspace{-13pt}
	\caption{The examples of discontinuous motion.}
	\label{fig:discontinuous}
\end{figure}

Most of the previous works focus on the motion of the objects in videos. They utilize the estimated flow maps~\cite{deepvoxelflow,superslomo}, kernels~\cite{long2016learning,adaconv,sepconv,lee2020adacof,ding2021cdfi}, or externally estimated optical flow maps~\cite{Niklaus2018CVPR,niklaus2020softmax,park2020bmbc,ABME,siyao2021deep} to place each object in the middle of its position on the adjacent frames. However, as personal broadcast and cloud gaming contents increase, many of the practical videos contain special objects which do not move continuously such as user interfaces, watermarks, logos, chatting windows and subtitles~(as the examples on Figure~\ref{fig:discontinuous}). Besides, these elements are received at the display devices as part of each frame, not as additional information. Therefore, many of the video enhancement frameworks, including VFI, should be improved to be robust to the discontinuous motions.

In this paper, our purpose is to expand the spectrum of motions to address efficiently both continuous and discontinuous ones, not focusing only on special videos with discontinuous motions. To achieve this goal, we propose three techniques to process the videos containing both types of motions. First, we propose novel data augmentation method called Figure-Text Mixing~(FTM) which consists of Figure Mixing~(FM) and Text Mixing~(TM). FM is an augmentation of fixed random figures and TM is an augmentation of discontinuity moving random texts. The networks can learn to be robust on both continuous and discontinuous motions using FTM without any additional train datasets. Second, we propose a lightweight module which estimates a map called discontinuity map~($D$-map), which determines whether the motion of each output pixel is continuous or discontinuous. When estimating the pixels in the discontinuous area, a pixel on the same location of one of the input frames is copied instead of predicting the interpolated value. To prove the versatility and adaptivity of our $D$-map estimation module, we apply it to various state-of-the-art VFI models instead of proposing our original VFI architecture. Lastly, if we utilize both FTM and $D$-map, it is possible to supervise the model by giving the ground-truth of $D$-map. Therefore, we propose an additional loss function to help our model estimate $D$-map easily.

We construct a special test set called Graphic Discontinuous Motion~(GDM) dataset to evaluate how our method and the competitive works deal with the discontinuous motions. Our approach shows significantly improved results compared to the other methods on GDM dataset. Moreover, our method outperforms those methods on the regular benchmarks that only contain continuous motions such as Vimeo90K test~\cite{xue2019video}, DAVIS~\cite{PontTusetarXiv2017} and UCF101~\cite{soomro2012ucf101} datasets. Our main contributions can be summarized as follows:

\begin{itemize}
\item \textit{New Data Augmentation Strategy.} We propose a new data augmentation strategy called FTM that, when applied to existing video datasets, makes models learn both continuous and discontinuous motions without any additional train datasets.
\item \textit{New Module \& Loss Function.} We propose a new module which can separate continuous and discontinuous motions. This module can be applied to many recent deep learning-based VFI architectures. We also propose a loss function to supervise the module.
\item \textit{Performance.} Applied to various state-of-the-art VFI models, our method achieves performance improvement on not only the dataset containing discontinuous motions, but also many of the other benchmarks with only continuous motions.
\end{itemize}

\section{Related Work}
Most existing video frame interpolation algorithms consist of two parts: motion estimation and motion compensation. Motion estimation modules estimate the pixel level correspondences between two consecutive frames to get motion information. Then motion compensation parts warp the frames according to the estimated motion. Recent video frame interpolation researches utilize deep neural networks~(DNN) to obtain high quality results in two ways.

One method is the end-to-end learning approach. Several works train their neural networks, which perform both motion estimation and compensation at the same time. Niklaus~\emph{et~al.}~\cite{adaconv} propose a  network that estimates big kernel weights for all pixels of input frames. Then they adaptively convolve the input frames with the estimated kernels to obtain the output frame. Since the large kernel size requires excessively many weights, Niklaus~\emph{et~al.}~\cite{sepconv} solve this problem by using separable kernels. On the other hand, Liu~\emph{et~al.}~\cite{deepvoxelflow} and Jiang~\emph{et~al.}~\cite{superslomo} propose the neural network that estimates dense flow map, which consists of the vectors directly pointing to the reference pixels. However, the above methods have limitations that the kernel-based ones cannot deal with the motion beyond the kernels, and the flow-based ones only refer to one pixel for each output pixel. To solve the problem, Lee~\emph{et~al.}~\cite{adaconv} combine the two methods using the deformable convolution~\cite{dai2017deformable}. Some approaches propose the neural networks that directly estimate the intermediate frames without motion compensation. Long~\emph{et~al.}~\cite{long2016learning} train a simple U-Net~\cite{ronneberger2015u} to estimate the intermediate frames, but the results tend to be blurry. Therefore, Choi~\emph{et~al.}~\cite{choi2020channel} propose a new architecture based on channel attention to obtain  sharper results. Recently, transformer-based algorithms~\cite{VFIformer, VFIT} have been proposed. Lu~\emph{et~al.}~\cite{VFIformer} identify a limitation of convolution operations which can only handle small motion and solve this with transformer architecture. Shi~\emph{et~al.}~\cite{VFIT} expand from  Lee~\emph{et~al.}~\cite{adaconv} to produce high quality result by adding transformer-based block.

The second method is the optical flow-based approach. Recently, many approaches to estimate high-quality optical flow maps have been introduced~\cite{ilg2017flownet,sun2018pwc,teed2020raft,park2020bmbc,ABME}. Therefore, several works make use of the optical flow maps as motion information and train additional networks for motion compensation or output frame refinement. Niklaus~\emph{et~al.}~\cite{Niklaus2018CVPR} utilize the context information extracted from ResNet-18~\cite{he2016deep} along with the optical flows and refine the warped frames using their own neural network based on GridNet~\cite{fourure2017residual}. Using pre-trained optical flow has a problem that the flow maps consist of the vectors starting from the input frames. However, the vectors starting from the output frame are necessary to warp the frames clearly. To deal with this problem, Bao~\emph{et~al.}~\cite{bao2019depth} use the depth maps obtained from the hourglass architecture-based mono depth estimation network~\cite{chen2016single} to invert the optical flow maps clearly. Niklaus~\emph{et~al.}~\cite{niklaus2020softmax} propose to combine all pixel values that are projected into the same locations using SoftMax function. Park~\emph{et~al.}~\cite{park2020bmbc} use the symmetric bilateral motions estimated as a linear to improve the quality of the motion. However, considering the limitation of handling the region where the constraint does not exist, Park~\emph{et~al.}~\cite{ABME} expands the symmetric to asymmetric motion for robust video frame interpolation. On the other hand, several algorithms are proposed in terms of efficiency. Kong~\emph{et~al.}~\cite{IFRNet} proposes warp and refine immediately with pyramid architecture and demonstrates that this approach is efficient compared with previous flow-based algorithms. Huang~\emph{et~al.}~\cite{RIFE} achieves real-time flow estimation by employing not pre-trained flow network but end-to-end convolution network.

There are some researches to expand the domain of videos to be interpolated. Xiangyu~\emph{et~al.}~\cite{xu2019quadratic} propose the quadratic video interpolation approach which utilizes four frames to cover not only linear motions, but also quadratic motions. However, they still cannot incorporate discontinuous motions. Lastly, Li~\emph{et~al.}~\cite{siyao2021deep} propose the network that can interpolate cartoon videos. However, they focus on the characteristics of cartoon images, not the motion of the videos. In this paper, we expand the video interpolation task to cover not only the natural motions, but also discontinuous transitions between the frames.

\section{Proposed Approach}

Given the two consecutive video frames $I_1 , I_2 \in \mathbb{R}^{H \times W \times C} $~(sometimes along with additional frames), recent deep learning-based VFI methods try to build a network $\mathcal{F}$ to estimate the intermediate frame $\hat{I}$. Omitting the additional frames, we can express the VFI frameworks as follows:

\begin{equation}
   \hat{I} = \mathcal{F}(I_1, I_2)
\end{equation}

\noindent However, most of the networks are designed to deal with only continuous motions. To make them robust to the discontinuous motions like in Figure~\ref{fig:discontinuous}, there are two main problems that need to be solved. First, they are mainly trained on the datasets containing only continuous motions such as Vimeo90K~\cite{xue2019video}. To solve this problem without collecting additional train dataset, we propose a data augmentation method called FTM, which is introduced in Section~\ref{sec:ftm}. Second, many of the architectures make use of motion information such as optical flow, which limits the models to considering only the continuous motions. To solve this problem, we propose a new module that predicts discontinuity map~($D$-map), which is introduced in Section~\ref{sec:dmap}. Applied to many existing networks, it makes them able to deal with both continuous and discontinuous motions better.

\begin{figure} 
	\begin{center}
		\includegraphics[width=1\linewidth]{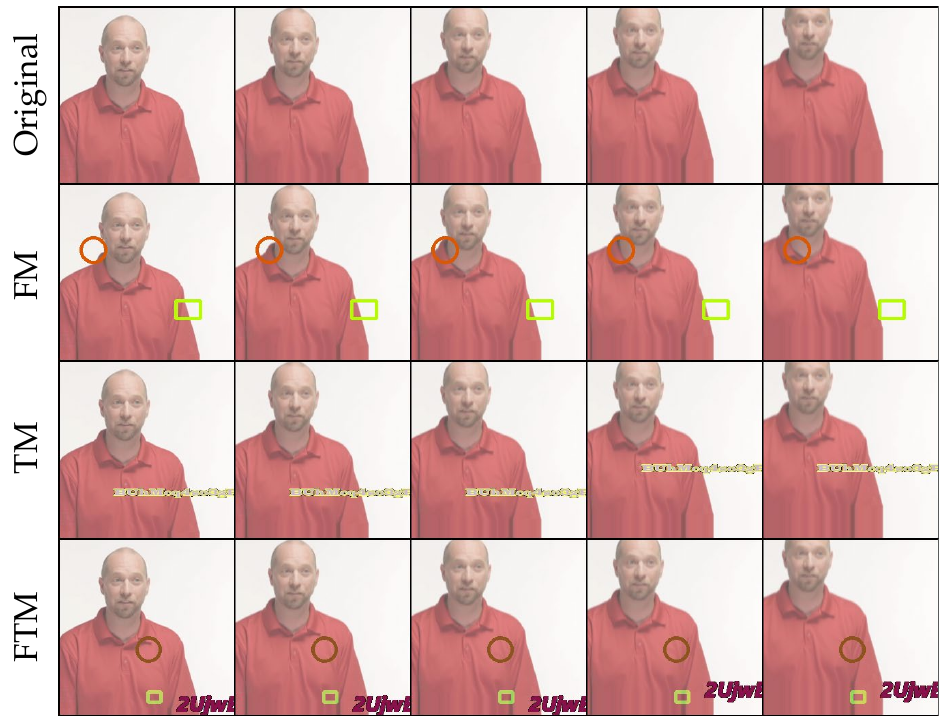}
	\end{center}
	\caption{Example of augmented training data. Figure-Text Mixing is applied randomly in units of sequence.}
	\label{fig:augmentation}
\end{figure}

\subsection{Figure-Text Mixing~(FTM)}
\label{sec:ftm}
Previous studies have only applied flip augmentation with spatial and temporal axes. However, this general augmentation is not sufficient alone to address various types of videos. We proposed a new data augmentation method for frame interpolation, called \textit{Figure-Text mixing}~(FTM), to handle general video frame interpolation. FTM consisted of two types of data augmentation methods: \textit{figure mixing}~(FM) and \textit{text mixing}~(TM). The discontinuity map, i.e., the results of our network, is the difference map between the original ground-truth frame and augmented ground-truth frame in the training stage. We can train the proposed network with the supervision of the discontinuity map by applying the data augmentation technique.

\noindent \textbf{Figure Mixing.} We added figures to the input frames to address the static objects in videos. The added figures had the same position and property on all input frames. We randomly applied the augmentation for the video sequence (see Figure \ref{fig:augmentation}). This technique can be the guideline where the discontinuous motion area exists. This method also maintained the edge of an object from collapsing, even if the object had continuous motion.

\noindent \textbf{Text Mixing.} There were many letters or sentences in videos, such as chatting and watermark. We included text on the video in the following four methods owing to the difference between the property of discontinuous and continuous motion areas: 1) the position of the text is static in the entire video, 2) where the text does not exist in the previous frame and appears in the future frame, 3) in the opposite of case 2), 4) the position of the text changes.
We set the method of FTM in which the ground-truth frame and the discontinuity map follow the previous frame augmentation in text mixing. Therefore, we applied data augmentation, as given in Figure \ref{fig:augmentation}. The first row of Figure \ref{fig:augmentation} is the case of 2), and the second row is the case of 4).

\begin{figure}
	\begin{center}
		\includegraphics[width=1.0\linewidth]{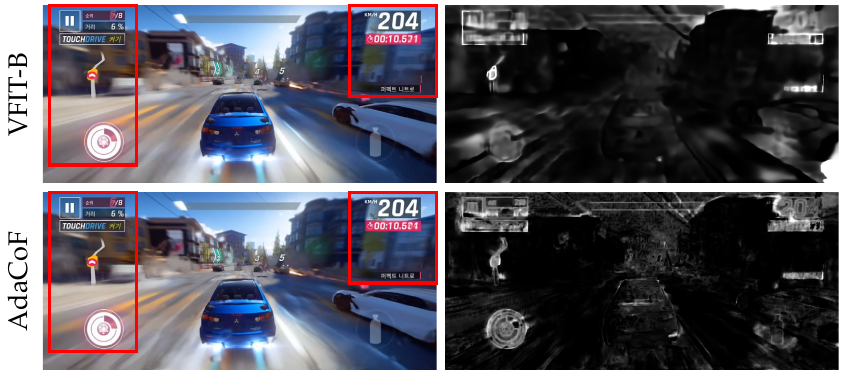}
	\end{center}
	\caption{Visualization of Discontinuity map~($D$-map) estimated for a sample. AdaCoF and VFIT-B are the baseline networks where we apply our methods. $D$-map commonly shows the user interface elements~(UI) that are fixed on their position and some digits that change dynamically~(highlighted in red boxes).}
\label{fig:dmap}
\end{figure}

\subsection{Discontinuity Map~($D$-map)}
\label{sec:dmap}
Some examples of discontinuous motions in Figure~\ref{fig:discontinuous} share an important property: interpolation of discontinuously moving object can be done by simply copying it onto one of the adjacent frames. However, this method may pose minor problems. First is the problem of determining which frame to copy between the front and back frames. The second is some samples where the \textit{copying strategy} does not work like in the fourth row of Figure~\ref{fig:discontinuous}. However, the problem of interpolating these cases is ill-posed fundamentally. For example, it is difficult to identify which digit should be generated when it changes frame-by-frame. Considering the visual quality, even for the above two cases, it is more appropriate to simply copy the previous or next frame than let the motion compensation model to solve this problem.

Therefore, we additionally estimate a 1-channel map called Discontinuity map~($D$-map) to distinguish the continuous and discontinuous motion areas. The pixels in continuous areas are estimated by the baseline VFI network, and those in discontinuous areas are copied from the previous frame. When $D \in (0,1)^{H \times W \times 1}$ represents a $D$-map, the proposed VFI process can be expressed as follows:

\begin{equation}
    \hat{I_c} = \mathcal{F}(I_1, I_2)
\end{equation}

\begin{equation}
    \hat{I}(\textbf{x}) = \hat{I_c}(\textbf{x}) \cdot (1 - D(\textbf{x})) + I_1(\textbf{x}) \cdot D(\textbf{x}),
\end{equation}

\noindent where $\hat{I_c}$ is the \textit{continuously} interpolated frame and $\textbf{x} \in [1,H] \times [1,W]$ represents the location. For example, we discovered that $D$-map commonly indicates UI elements or digits which have discontinuous motion regardless of baseline networks in Figure~\ref{fig:dmap}. More examples of $D$-map are shown in supplementary material.

\noindent \textbf{Discontinuity map estimation.} Given a conventional VFI network $\mathcal{F}$ as a baseline, $D$-map is estimated by a lightweight CNN-based module. The input of this module is a feature map obtained from a specific layer of $\mathcal{F}$. We select the most appropriate layer for each network and the details are presented in supplementary material. In addition, we increase the number of input frames of the baseline networks to four because the \textit{continuity} and \textit{discontinuity} cannot be defined with only two frames. For fair comparison with the previous methods, we also compare the 2-frame versions in Section~\ref{sec:exp}. Finally, the modified network~(4 frame input and $D$-map estimation module-added) is trained with the loss functions that are introduced in Section~\ref{sec:loss}~(with FTM applied as well).

\begin{figure}
	\begin{center}
		\includegraphics[width=1.0\linewidth]{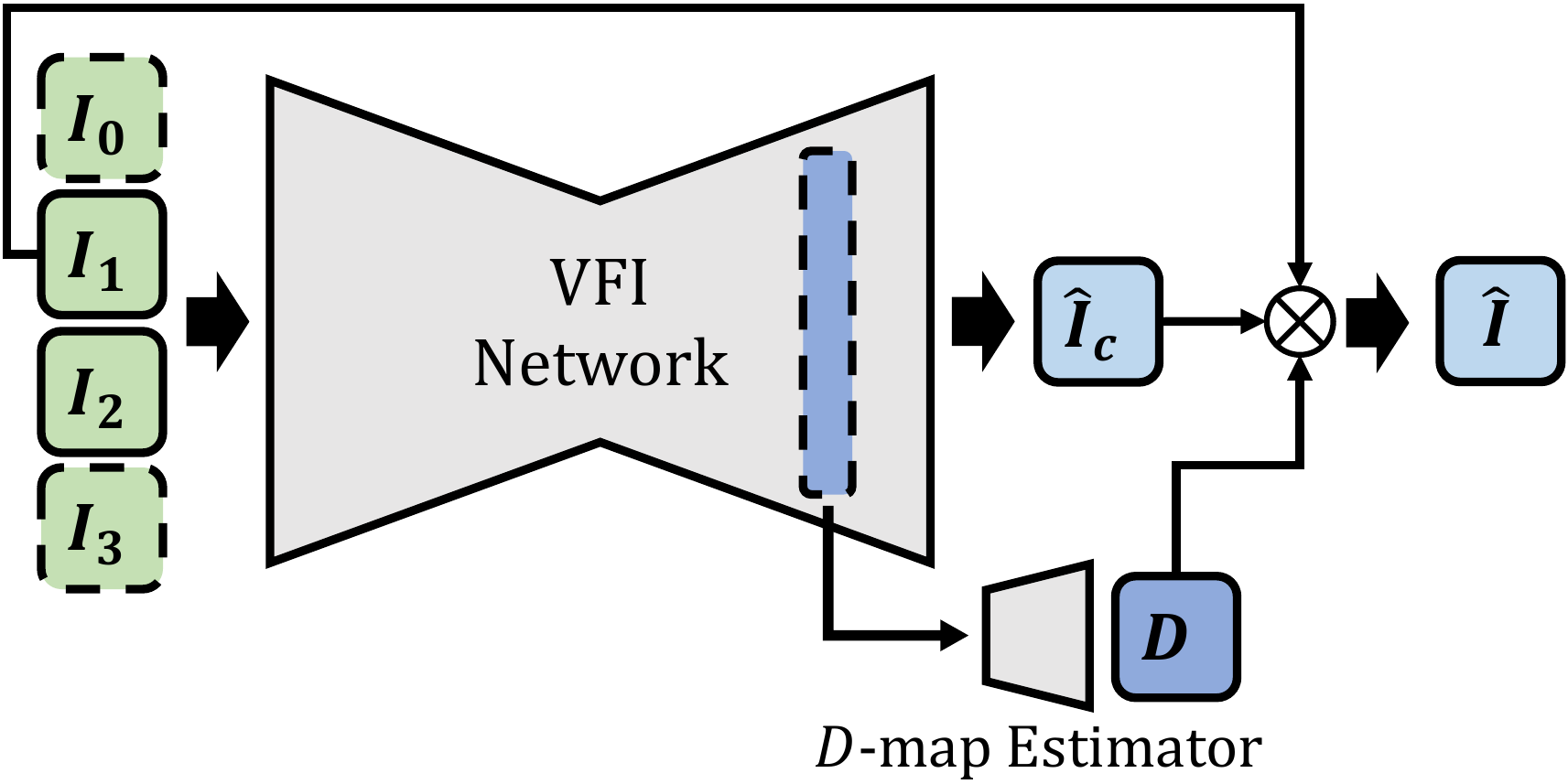}
	\end{center}
	\caption{The architecture of our method. VFI network can be various state-of-the-art models. Our modifications from the baseline are the number of frames and $D$-map estimator.}
\label{fig:architecture}
\end{figure}

\begin{table*}
\footnotesize
	\begin{center}
			\begin{tabular}{lcccccccccccc}
				\toprule
				& \multicolumn{3}{c}{Vimeo-90k} & \multicolumn{3}{c}{UCF101} & \multicolumn{3}{c}{DAVIS} & \multicolumn{3}{c}{GDM} \\
				\cmidrule(l{5pt}r{5pt}){2-4} \cmidrule(l{5pt}r{5pt}){5-7} \cmidrule(l{5pt}r{5pt}){8-10} \cmidrule(l{5pt}r{5pt}){11-13}
				& PSNR  & SSIM  & LPIPS & PSNR  & SSIM  & LPIPS  & PSNR  & SSIM  & LPIPS & PSNR  & SSIM  & LPIPS \\
				\midrule
				DVF~\cite{deepvoxelflow} & 32.792 & 0.9359 & 0.0395 & 32.122 & 0.9366 & \textbf{0.0356} & 25.596 & \textbf{0.8084} & 0.1452 & 28.709 & 0.9118 & 0.0753\\			
				DVF-\textit{FTM} & \textbf{32.833} & \textbf{0.9361} & \textbf{0.0395} & \textbf{32.162} & \textbf{0.9369} & 0.0360 & \textbf{25.605} & 0.8078 & \textbf{0.1449} & \textbf{28.984} & \textbf{0.9134} & \textbf{0.0728}\\	
				\midrule
				SepConv~\cite{sepconv} & 32.889 & 0.9306 & 0.0681 & 32.993 & 0.9416 & 0.0429 & \textbf{25.952} & \textbf{0.8108} & 0.2209 & 29.667 & 0.9197 & 0.0870\\			
				SepConv-\textit{FTM} & \textbf{33.064} & \textbf{0.9325} & \textbf{0.0665} & \textbf{33.157} & \textbf{0.9428} & \textbf{0.0418} & 25.873 & 0.8103 & \textbf{0.2188} & \textbf{29.914} & \textbf{0.9235} & \textbf{0.0797}\\				
				\midrule
				AdaCoF~\cite{lee2020adacof} & 34.103 & 0.9459 & 0.0427 & 33.320 & \textbf{0.9438} & 0.0353 & \textbf{26.791} & \textbf{0.8353} & 0.1643 & 29.980 & 0.9227 & 0.0803\\
				AdaCoF-\textit{FTM} & \textbf{34.167} & \textbf{0.9497} & \textbf{0.0396} & \textbf{33.333} & 0.9437 & \textbf{0.0345} & 26.719 & 0.8342 & \textbf{0.1561} & \textbf{30.064} & \textbf{0.9278} & \textbf{0.0728}\\
				\midrule
				CAIN~\cite{choi2020channel} & 34.699 & 0.9514 & 0.0421 & 33.306 & 0.9444 & 0.0373 & \textbf{27.449} & 0.8511 & \textbf{0.1855} & 30.238 & 0.9284 & 0.0807\\			
				CAIN-\textit{FTM} & \textbf{34.987} & \textbf{0.9537} & \textbf{0.0414} & \textbf{33.448} & \textbf{0.9453} & \textbf{0.0363} & 27.427 & \textbf{0.8528} & 0.1879 & \textbf{30.429} & \textbf{0.9328} & \textbf{0.0776}\\	
				\midrule
				VFIT-B~\cite{VFIT} & \textbf{36.743} & \textbf{0.9638} & \textbf{0.0318} & 33.769 & 0.9472 & \textbf{0.0363} & 28.090 & \textbf{0.8640} & \textbf{0.1442} & 30.019 & 0.9280 & 0.0736\\			
				VFIT-B-\textit{FTM} & 36.695 & 0.9633 & 0.0322 & \textbf{33.821} & \textbf{0.9474} & 0.0377 & \textbf{28.140} & 0.8627 & 0.1506 & \textbf{30.473} & \textbf{0.9321} & \textbf{0.0693}\\	
				\bottomrule
		\end{tabular}
		\caption{Effect of the Figure-Text Mixing~(FTM) augmentation. The label '-\textit{FTM}' means the model trained with FTM applied. The \textbf{bold highlights} mean that the performances are improved by FTM.}
		\label{tbl:ftm}
	\end{center}
\end{table*}

\subsection{Objective Functions}
\label{sec:loss}
\noindent\textbf{Loss Function.} First, we reduced the difference between the model output $I_{out}$ and ground truth $I_{gt}$. We used $\ell_1$ norm for the loss as follows:
\vspace{-0.1cm}
\begin{equation}
\mathcal{L}_1 = \|I_{out} - I_{gt}\|_1~.
\end{equation}

\noindent We use the Charbonnier Function $\Phi(x)=(x^2 + \epsilon^2)^{1/2}$ for optimizing $\ell_1$ distance, where $\epsilon=0.001$ by following Liu~\emph{et~al.}~\cite{deepvoxelflow}.

\noindent \textbf{Discontinuity map Supervision.} When both FTM and $D$-map are applied in training stage, the ground truth of $D$-map $D_{gt}$ can be obtained by simply getting the areas of added figures and texts. Therefore, we can provide the $D$-map supervision to the network to learn the location of discontinuous motion areas. We use $\ell_1$ loss between $D$ and $D_{gt}$ as follows:

\begin{equation}
\mathcal{L}_{D} = \|D - D_{gt}\|_1~.
\end{equation}

\noindent We find that just adding the two losses with same rate works well. Therefore, the total loss for training is as follows:
\begin{equation}
\mathcal{L}_{total} = \mathcal{L}_1 + \mathcal{L}_{D}~.
\end{equation}

\section{Experiments}
\label{sec:exp}
\subsection{Experimental Settings}
\label{sec:setting}
\noindent \textbf{Training Settings.} We use Vimeo90K~\cite{xue2019video} septuplet train dataset for training, which consists of 91,701 sequences of seven $448 \times 256$ frames. For the data augmentations, we randomly crop the $256 \times 256$ patches and flip them horizontally, vertically, and temporally. We then add the proposed data augmentation~(FTM). For the other options such as learning rates, scheduling, batch sizes, and the optimizers, we follow the respective papers where each baseline network is proposed. Note that the proposed methods are all that can be applied to various baseline models.

\noindent \textbf{Evaluation Setting.} For the evaluation, we select some recent VFI networks each representing the warping-based~\cite{lee2020adacof}, direct prediction-based~\cite{choi2020channel}, and transformer-based~\cite{VFIT} architecture. Then we observe the performance changes as the three proposed ideas~(FTM, $D$-map, and $\mathcal{L}_{D}$) are applied. We select three test benchmarks that are commonly used: Vimeo90K~\cite{xue2019video} test dataset, UCF101~\cite{soomro2012ucf101}, DAVIS test-dev dataset~\cite{PontTusetarXiv2017}. In addition, we construct a new test set called Graphic Discontinuous Motion~(GDM) dataset which consists of high resolution videos of game scenes with abundant discontinuous motions. There are some more details of GDM dataset in the supplementary material. We evaluate each algorithm by comparing PSNR~(Peak Signal-to-Noise Ratio), SSIM~(Structural Similarity)~\cite{wang2004image}, and LPIPS~(Learned Perceptual Image Patch Similarity)~\cite{zhang2018unreasonable} for all test datasets.

\begin{table*}
\footnotesize
	\begin{center}
			\begin{tabular}{lcccccccccccc}
				\toprule
				& \multicolumn{3}{c}{Vimeo-90k} & \multicolumn{3}{c}{UCF101} & \multicolumn{3}{c}{DAVIS} & \multicolumn{3}{c}{GDM} \\
				\cmidrule(l{5pt}r{5pt}){2-4} \cmidrule(l{5pt}r{5pt}){5-7} \cmidrule(l{5pt}r{5pt}){8-10} \cmidrule(l{5pt}r{5pt}){11-13}
				& PSNR  & SSIM  & LPIPS & PSNR  & SSIM  & LPIPS  & PSNR  & SSIM  & LPIPS & PSNR  & SSIM  & LPIPS \\
				\midrule
				AdaCoF~(2)~\cite{lee2020adacof} & 34.103 & 0.9459 & 0.0427 & 33.320 & 0.9438 & 0.0353 & 26.791 & 0.8353 & 0.1643 & 29.980 & 0.9227 & 0.0803\\
				AdaCoF-$D$~(2) & 34.385 & 0.9483 & 0.0412 & 33.355 & 0.9439 & 0.0358 & 26.754 & 0.8356 & 0.1605 & 30.097 & 0.9260 & 0.0787\\
				AdaCoF-$D$~(4) & 35.240 & 0.9534 & \textbf{0.0384} & 33.595 & 0.9452 & 0.0353 & 27.171 & \textbf{0.8425} & 0.1635 & 30.212 & 0.9255 & 0.0793\\
				AdaCoF-$D$-$\mathcal{L}_D$~(4) & \textbf{35.267} & \textbf{0.9535} & 0.0385 & \textbf{33.611} & \textbf{0.9453} & \textbf{0.0350} & \textbf{27.226} & 0.8424 & \textbf{0.1597} & \textbf{30.242} & \textbf{0.9296} & \textbf{0.0760}\\
				\midrule
				CAIN~(2)~\cite{choi2020channel} & 34.699 & 0.9514 & 0.0421 & 33.306 & 0.9444 & \textbf{0.0373} & 27.449 & 0.8511 & 0.1855 & 30.238 & 0.9284 & 0.0807\\			
				CAIN-$D$~(2) & 34.546 & 0.9500 & 0.0455 & 33.319 & 0.9445 & 0.0375 & 27.301 & 0.8499 & 0.1927 & 30.616 & 0.9314 & 0.0792\\	
				CAIN-$D$~(4) & \textbf{35.540} & \textbf{0.9560} & \textbf{0.0405} & \textbf{33.618} & \textbf{0.9461} & 0.0376 & \textbf{28.021} & \textbf{0.8612} & \textbf{0.1852} & \textbf{31.011} & \textbf{0.9375} & \textbf{0.0664}\\	
				CAIN-$D$-$\mathcal{L}_D$~(4) & 35.169 & 0.9526 & 0.0451 & 33.589 & 0.9456 & 0.0390 & 27.897 & 0.8571 & 0.2019 & 30.936 & 0.9293 & 0.0703\\
				\midrule
				VFIT-B~(4)~\cite{VFIT} & \textbf{36.743} & \textbf{0.9638} & \textbf{0.0318} & 33.769 & 0.9472 & \textbf{0.0363} & \textbf{28.090} & \textbf{0.8640} & \textbf{0.1442} & 30.019 & 0.9280 & 0.0736\\			
				VFIT-B-$D$~(4) & 36.650 & 0.9634 & 0.0320 & 33.819 & 0.9474 & 0.0367 & 28.026 & 0.8621 & 0.1507 & \textbf{30.965} & \textbf{0.9381} & 0.0652\\	
				VFIT-B-$D$-$\mathcal{L}_D$~(4) & 36.671 & 0.9631 & 0.0324 & \textbf{33.823} & \textbf{0.9475} & 0.0370 & 28.056 & 0.8625 & 0.1507 & 30.921 & 0.9371 & \textbf{0.0645}\\	
				\bottomrule
		\end{tabular}
		\caption{Effect of applying $D$-map and $\mathcal{L}_D$. '-$D$' and '-$\mathcal{L}_D$' indicate the versions applying $D$-map and $\mathcal{L}_D$ each. The labels '(2)' and '(4)' mean the number of input frames. The \textbf{bold highlights} imply that the performances are the best among the versions sharing the same baseline.}
		\label{tbl:dmap}
	\end{center}
\end{table*}

\begin{table*}[t]
\footnotesize
	\begin{center}
			\begin{tabular}{lcccccccccccc}
				\toprule
				& \multicolumn{3}{c}{Vimeo-90k} & \multicolumn{3}{c}{UCF101} & \multicolumn{3}{c}{DAVIS} & \multicolumn{3}{c}{GDM} \\
				\cmidrule(l{5pt}r{5pt}){2-4} \cmidrule(l{5pt}r{5pt}){5-7} \cmidrule(l{5pt}r{5pt}){8-10} \cmidrule(l{5pt}r{5pt}){11-13}
				& PSNR  & SSIM  & LPIPS & PSNR  & SSIM  & LPIPS  & PSNR  & SSIM  & LPIPS & PSNR  & SSIM  & LPIPS \\
				\midrule
				DVF~(2)~\cite{deepvoxelflow} & 32.792 & 0.9359 & 0.0395 & 32.333 & 0.9397 & 0.0340 & 24.087 & 0.7852 & 0.1588 & 28.709 & 0.9118 & 0.0753\\
				SuperSlomo~(2)~\cite{superslomo}  & 30.812 & 0.9291 & 0.0482 & 28.500 & 0.9228 & 0.0564 & 26.259 & 0.8303 & 0.1206 & 27.651 & 0.8911 & 0.1117\\
				SepConv-$\mathcal{L}_1$~(2)~\cite{sepconv}  & 33.729 & 0.9454 & 0.0335 & 33.075 & 0.9419 & 0.0333 & 26.550 & 0.8376 & 0.1478 & 29.696 & 0.9082 & 0.1037 \\
				AdaCoF~(2)~\cite{lee2020adacof}  & 34.103 & 0.9459 & 0.0427 & 33.320 & 0.9438 & 0.0353 & 26.791 & 0.8353 & 0.1643 & 29.980 & 0.9227 & 0.0803 \\
				Softsplat-$\mathcal{L}_1$~(2)~\cite{niklaus2020softmax}  & 33.723 & 0.9452 & 0.0336 & 33.112 & 0.9419 & 0.0332 & 26.542 & 0.8376 & 0.1479 & 29.667 & 0.9086 & 0.1039 \\
				CAIN~(2)~\cite{choi2020channel} & 34.699 & 0.9514 & 0.0421 & 33.306 & 0.9444 & 0.0373 & 27.449 & 0.8511 & 0.1855 & 30.238 & 0.9284 & 0.0807\\
				ABME~(2)~\cite{ABME} & 35.846 & 0.9584 & 0.0309 & 33.542 & 0.9458 & 0.0383 & 27.661 & 0.8601 & 0.1320 & 29.472 & 0.9209 & 0.0958 \\
				RIFE~(2)~\cite{RIFE} & 34.048 & 0.9449 & \textbf{\textcolor{red}{0.0233}} & 33.184 & 0.9412 & \textbf{\textcolor{red}{0.0284}} & 27.246 & 0.8471 & \textbf{\textcolor{red}{0.0925}} & 30.085 & 0.9088 & 0.0801 \\
				IFRNet~(2)~\cite{IFRNet} & 35.837 & 0.9597 & \textbf{\textcolor{blue}{0.0274}} & 33.451 & 0.9450 & \textbf{\textcolor{blue}{0.0330}} & 27.467 & 0.8596 & \textbf{\textcolor{blue}{0.1261}} & 30.239 & 0.9277 & \textbf{\textcolor{blue}{0.0706}} \\
				VFIT-B~(4)~\cite{VFIT} & \textbf{\textcolor{red}{36.963}} & \textbf{\textcolor{red}{0.9649}} & 0.0304 & \textbf{\textcolor{red}{33.837}} & \textbf{\textcolor{blue}{0.9474}} & 0.0367 & \textbf{\textcolor{red}{28.153}} & \textbf{\textcolor{red}{0.8652}} & 0.1440 & 30.217 & 0.9274 & 0.0760 \\
				\midrule
				CAIN-$D$~(2) & 34.546 & 0.9500 & 0.0455 & 33.319 & 0.9445 & 0.0375 & 27.301 & 0.8499 & 0.1927 & \textbf{\textcolor{blue}{30.616}} & \textbf{\textcolor{blue}{0.9314}} & 0.0792\\
				VFIT-B-$D$-$\mathcal{L}_D$~(4) & \textbf{\textcolor{blue}{36.671}} & \textbf{\textcolor{blue}{0.9631}} & 0.0324 & \textbf{\textcolor{blue}{33.823}} & \textbf{\textcolor{red}{0.9475}} & 0.0370 & \textbf{\textcolor{blue}{28.056}} & \textbf{\textcolor{blue}{0.8625}} & 0.1507 & \textbf{\textcolor{red}{30.921}} & \textbf{\textcolor{red}{0.9371}} & \textbf{\textcolor{red}{0.0645}}\\
				\bottomrule
		\end{tabular}
		\caption{Quantitative evaluation on three datasets. The labels '(2)' and '(4)' indicate the number of input frames. The \textbf{\textcolor{red}{red highlights}} define the best,  \textbf{\textcolor{blue}{blue highlights}} second best.}
  \vspace{-.3cm}
    \label{tbl:main result}
	\end{center}
\end{table*}

\subsection{Ablation Study}
\label{ablation}
\noindent \textbf{Figure-Text Mixing~(FTM).} To figure out the effectiveness of FTM, we first train some recent VFI networks~(AdaCoF~\cite{lee2020adacof}, DVF~\cite{deepvoxelflow}, SepConv~\cite{sepconv}, CAIN~\cite{choi2020channel}, and VFIT-B~\cite{VFIT}) following their own training strategy. Then we train them again in the same ways with FTM applied. Table~\ref{tbl:ftm} presents that FTM significantly improves the interpolation performance especially in GDM dataset. FTM is also effective on the conventional test benchmarks with only continuous motions such as Viemo90k, UCF101, and DAVIS in many cases.

\noindent \textbf{Discontinuity Map.} As mentioned in Section~\ref{sec:setting}, we apply $D$-map to the three recent VFI networks: AdaCoF~\cite{lee2020adacof}, CAIN~\cite{choi2020channel}, and VFIT-B~\cite{VFIT}. Then we train three or four versions of each network. First is the baseline network, second is the $D$-map applied version and the third is the $D$-map applied one with the supervision loss $\mathcal{L}_D$. Except for VFIT-B which requires four input frames, we additionally train the versions with two input frames for fair comparison~(note that $D$-map applied versions basically require four frames). Table~\ref{tbl:dmap} presents how $D$-map and $\mathcal{L}_D$ affect the performance. In short, applying $D$-map estimation module significantly improves the interpolation quality on GDM dataset. Moreover, some results show improved performance for the other datasets as well, and relatively small amounts of degradation gaps in some cases. For the models with only two input frames, AdaCoF-$D$~(2) illustrates better performance than the baseline. CAIN-$D$~(2) sometimes reveals lower quality on Vimeo 90K and DAVIS dataset, while it shows improvement on GDM dataset. Considering that our objective is to make VFI models robust to discontinuous motions without significantly degrading performance over continuous motions, the ideas clearly achieve our intention. In addition, we can also confirm that $\mathcal{L}_D$ is more effective on AdaCoF, compared to the other baselines.

\subsection{Quantitative Results}
\label{sec:Quantity}
We evaluate our network with several previous algorithms including DVF~\cite{deepvoxelflow}, SuperSlomo~\cite{superslomo}, SepConv~\cite{sepconv}, AdaCoF~\cite{lee2020adacof}, Softsplat~\cite{niklaus2020softmax}, CAIN~\cite{choi2020channel}, ABME~\cite{ABME}, RIFE~\cite{RIFE}, IFRNet~\cite{IFRNet}, and VFIT-B~\cite{VFIT}.
Since we have many possible versions, we select two models for comparison: CAIN-$D$~(2) representing two-input-frames model and VFIT-B + $D$-map-4 + $\mathcal{L}_D$ representing four-input-frames model. As shown in Table~\ref{tbl:main result}, our methods outperform the previous algorithms on GDM dataset for three metrics by a high margin~(note that the previous algorithms are pretrained models provided and our two networks are re-trained with the proposed method). Especially, despite the lower performance of CAIN baseline compared to the other networks, CAIN-$D$~(2) achieves second best performance on GDM dataset. These results clearly show both the failures of the previous methods and the robustness of our approach for discontinuous motion. Moreover, our methods still show competitive performance among the state-of-the-art VFI networks even on continuous motions.

\vspace{.9cm}
Table~\ref{tbl:ftm} shows that FTM can be used for training any algorithms and leads high performance for discontinuous motion. We can obtain better performance on GDM dataset for five algorithms without using extra training datasets.
Table~\ref{tbl:dmap} shows that our ideas clearly improve the existing models. These results proves that our module can apply to any previous models and successfully improve the performance of baseline models on GDM dataset. Our methods achieve high performances on GDM dataset with slightly lower or higher quality on other three datasets.

\begin{figure*}[t]
	\begin{center}
		\includegraphics[width=0.99\linewidth]{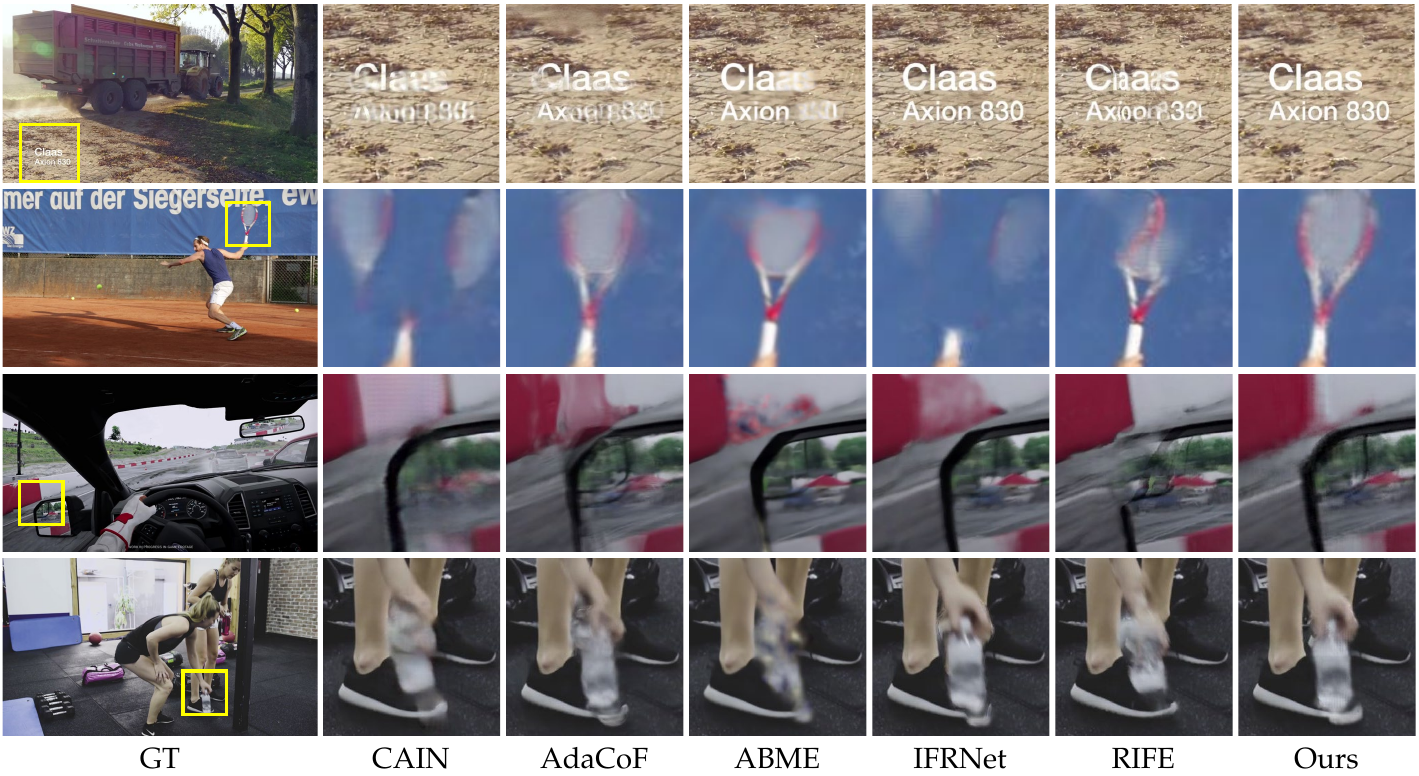}
	\end{center}
	\vspace{-0.5cm}
	\caption{Qualitative results on DAVIS test-dev dataset.}
	\label{fig:DAVIS result}
\end{figure*}

\subsection{Qualitative Results}
\label{sec:Quality}
\begin{figure*}[ht]
	\setlength{\belowcaptionskip}{-10pt}
	\begin{center}
		\includegraphics[width=0.99\linewidth]{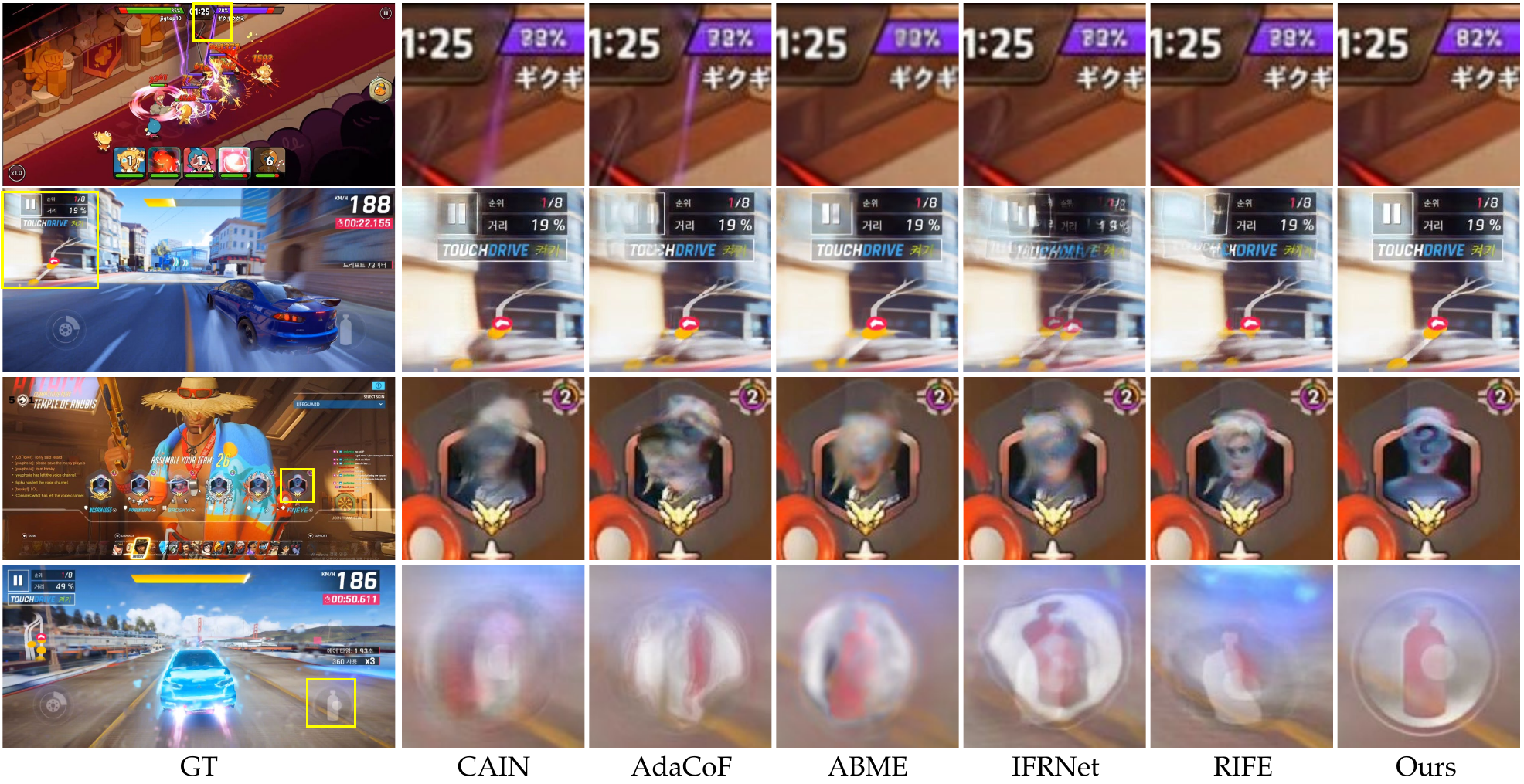}
	\end{center}
	\vspace{-0.5cm}
	\caption{Visual comparison of with discontinuous motion.}
	\label{fig:GDM result}
\end{figure*}

\begin{figure*}[t]
	\begin{center}
		\includegraphics[width=0.95\linewidth]{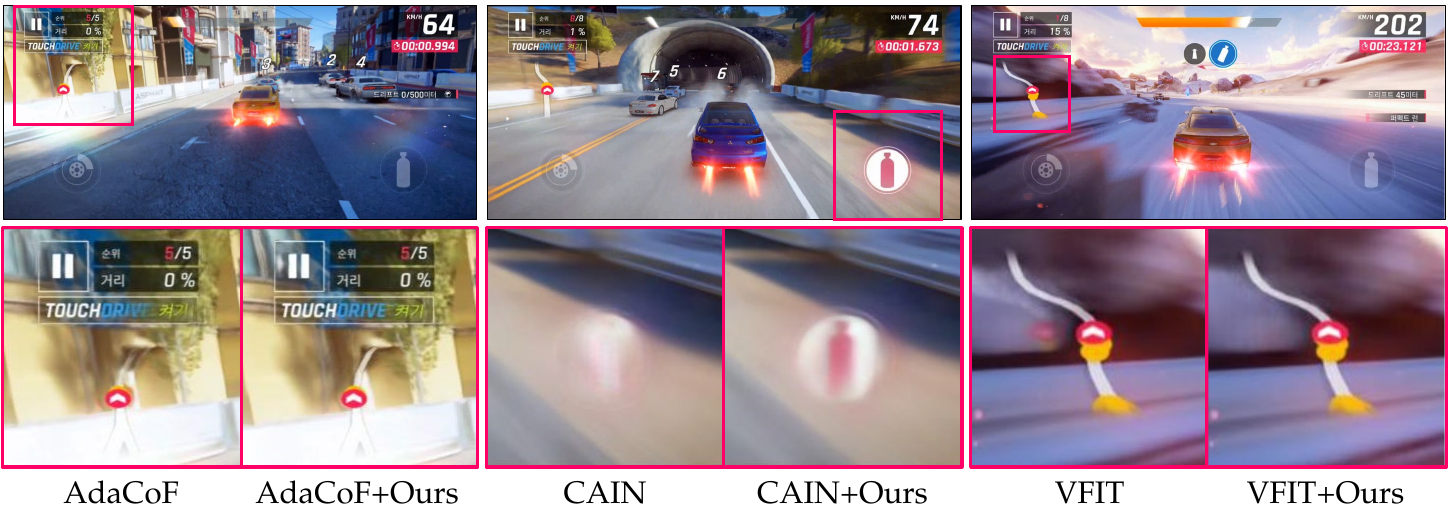}
	\end{center}
	\vspace{-0.5cm}
	\caption{Visual comparison between baseline algorithms and re-training with our methods.}
	\label{fig:FTM}
\end{figure*}

We apply our methods on various types of videos to demonstrate the effectiveness of our ideas. We visually compare the results on two types of datasets each containing continuous and discontinuous motions. Since there are multiple versions of the proposed method, we select the results from VFIT-B-$D$-$\mathcal{L}_D$ in Figures~\ref{fig:DAVIS result}, \ref{fig:GDM result}, \ref{fig:FTM} as \textit{Ours}.

\noindent \textbf{Continuous motion.}  We select the DAVIS test-dev dataset~\cite{PontTusetarXiv2017} to compare the results on continuous motions, generally addressed in previous studies. We compare our method with CAIN~\cite{choi2020channel}, AdaCoF~\cite{lee2020adacof}, ABME~\cite{ABME}, IFRNet~\cite{IFRNet}, and RIFE~\cite{RIFE}. As shown in Figure~\ref{fig:DAVIS result}, the other algorithms suffer from afterimage effects for large motion~(see 2nd~4th row of Figure~\ref{fig:DAVIS result}). On the other hand, our method shows relatively clear results. In the case of the sequence that has both continuous and discontinuous motions~(1st row of Figure~\ref{fig:DAVIS result}), the texts which need to maintain their rigid shape are distorted by the previous methods while our method show relatively clear results. In conclusion, compared to the other approaches, our method not only maintains the quality for continuous domain, but also produces sharp and clear results for large motion by maintaining the structure of the objects.

\noindent \textbf{Discontinuous motion.} We conduct two types of comparisons to demonstrate the effectiveness on discontinuous motion: comparison with previous algorithms on discontinuous video sequences from GDM dataset in Figure~\ref{fig:GDM result}, and comparison between baseline networks and ones with our methods applied in Figure~\ref{fig:FTM}. In Figure~\ref{fig:GDM result}, there are various types of discontinuous motions that were illustrated in Figure~\ref{fig:discontinuous}. We focus on two issues from the results of previous methods: 1) they sometimes fail to distinguish between continuous and discontinuous motion area, and 2) they have difficulty maintaining the structure of the discontinuously moving object, while they perform well in continuous motion area. Especially for the fixed objects, although all the VFI models need to do is just copying them from the previous frame, they sometimes suffer from some artifacts, while our method shows the clear shape compared with previous algorithms~(See 2nd and 4th row of Figure~\ref{fig:GDM result}). Similarly, the previous algorithms produce interpolated or inaccurate image for changing numbers and suddenly appearing objects~(See 1st and 3rd row of Figure~\ref{fig:GDM result}). However, our method shows better results by recognizing the momentarily changing scene. Figure~\ref{fig:GDM result} proves the limitations of previous algorithms on discontinuous motions and shows that our ideas can effectively cover both continuous and discontinuous domains. 

Figure~\ref{fig:FTM} shows the flexibility and effectiveness of our methods by revealing each baseline networks with applied our ideas. +Ours indicates applying FTM, $D$-map and $\mathcal{L}_D$ to each baseline network. The results from baseline algorithms suffer from some artifacts, but applying our ideas leads to clear and sharper results on discontinuous motions. Figure~\ref{fig:FTM} clearly shows that our methods can help the existing VFI models to be robust to the discontinuous motions and can be applied to various baselines.

\section{Limitations}
While FTM and $D$-map approaches can cover most cases of discontinuous motions, there are many other cases beyond that coverage. The main reason of the limitations is that our definition of discontinuous motion and the proposed ideas are somewhat naive, which make the solution sub-optimal. However, it is true that our approach can make the existing VFI models significantly more robust. Therefore, we hope some further researchers focus on more fundamental natures of discontinuous motions and propose more optimal solutions.

\section{Conclusion}
In this paper, we present the novel network for general video frame interpolation to address both continuous and discontinuous motion areas. We also propose a new data augmentation, named figure-text mixing~(FTM), to handle discontinuous motion areas and to overcome large motion. The evaluation shows state-of-the-art results for all kinds of datasets and indicators. These results imply that our network performs well for the general motion area without using extra datasets. We also demonstrated that the proposed architecture and data augmentation are effective solutions to handle general motion in various domains in ablation studies.

\noindent \textbf{Acknowledgements.}  This research was supported by R\&D program for Advanced Integrated-intelligence for Identification (AIID) through the National Research Foundation of KOREA(NRF) funded by Ministry of Science and ICT (NRF-2018M3E3A1057289) and the KIST Institutional Program(Project No.2E32283-23-064).

\clearpage
\clearpage

{\small
\bibliographystyle{ieee_fullname}
\bibliography{egbib}

\begin{thebibliography}{10}\itemsep=-1pt

\bibitem{bao2019depth}
Wenbo Bao, Wei-Sheng Lai, Chao Ma, Xiaoyun Zhang, Zhiyong Gao, and Ming-Hsuan
  Yang.
\newblock Depth-aware video frame interpolation.
\newblock In {\em Proceedings of the IEEE Conference on Computer Vision and
  Pattern Recognition}, pages 3703--3712, 2019.

\bibitem{bao2018high}
Wenbo Bao, Xiaoyun Zhang, Li Chen, Lianghui Ding, and Zhiyong Gao.
\newblock High-order model and dynamic filtering for frame rate up-conversion.
\newblock {\em IEEE Transactions on Image Processing}, 27(8):3813--3826, 2018.

\bibitem{chen2016single}
Weifeng Chen, Zhao Fu, Dawei Yang, and Jia Deng.
\newblock Single-image depth perception in the wild.
\newblock In {\em Advances in neural information processing systems}, pages
  730--738, 2016.

\bibitem{choi2020channel}
Myungsub Choi, Heewon Kim, Bohyung Han, Ning Xu, and Kyoung~Mu Lee.
\newblock Channel attention is all you need for video frame interpolation.
\newblock In {\em Proceedings of the AAAI Conference on Artificial
  Intelligence}, volume~34, pages 10663--10671, 2020.

\bibitem{dai2017deformable}
Jifeng Dai, Haozhi Qi, Yuwen Xiong, Yi Li, Guodong Zhang, Han Hu, and Yichen
  Wei.
\newblock Deformable convolutional networks.
\newblock In {\em The IEEE International Conference on Computer Vision (ICCV)},
  Oct 2017.

\bibitem{ding2021cdfi}
Tianyu Ding, Luming Liang, Zhihui Zhu, and Ilya Zharkov.
\newblock Cdfi: Compression-driven network design for frame interpolation.
\newblock In {\em Proceedings of the IEEE/CVF Conference on Computer Vision and
  Pattern Recognition}, pages 8001--8011, 2021.

\bibitem{flynn2016deepstereo}
John Flynn, Ivan Neulander, James Philbin, and Noah Snavely.
\newblock Deepstereo: Learning to predict new views from the world's imagery.
\newblock In {\em Proceedings of the IEEE conference on computer vision and
  pattern recognition}, pages 5515--5524, 2016.

\bibitem{fourure2017residual}
Damien Fourure, R{\'e}mi Emonet, Elisa Fromont, Damien Muselet, Alain Tremeau,
  and Christian Wolf.
\newblock Residual conv-deconv grid network for semantic segmentation.
\newblock {\em arXiv preprint arXiv:1707.07958}, 2017.

\bibitem{he2016deep}
Kaiming He, Xiangyu Zhang, Shaoqing Ren, and Jian Sun.
\newblock Deep residual learning for image recognition.
\newblock In {\em The IEEE Conference on Computer Vision and Pattern
  Recognition (CVPR)}, June 2016.

\bibitem{RIFE}
Zhewei Huang, Tianyuan Zhang, Wen Heng, Boxin Shi, and Shuchang Zhou.
\newblock Rife: Real-time intermediate flow estimation for video frame
  interpolation.
\newblock {\em arXiv preprint arXiv:2011.06294}, 2020.

\bibitem{ilg2017flownet}
Eddy Ilg, Nikolaus Mayer, Tonmoy Saikia, Margret Keuper, Alexey Dosovitskiy,
  and Thomas Brox.
\newblock Flownet 2.0: Evolution of optical flow estimation with deep networks.
\newblock In {\em IEEE conference on computer vision and pattern recognition
  (CVPR)}, volume~2, page~6, 2017.

\bibitem{superslomo}
Huaizu Jiang, Deqing Sun, Varun Jampani, Ming-Hsuan Yang, Erik Learned-Miller,
  and Jan Kautz.
\newblock Super slomo: High quality estimation of multiple intermediate frames
  for video interpolation.
\newblock In {\em The IEEE Conference on Computer Vision and Pattern
  Recognition (CVPR)}, June 2018.

\bibitem{kalantari2016learning}
Nima~Khademi Kalantari, Ting-Chun Wang, and Ravi Ramamoorthi.
\newblock Learning-based view synthesis for light field cameras.
\newblock {\em ACM Transactions on Graphics (TOG)}, 35(6):1--10, 2016.

\bibitem{IFRNet}
Lingtong Kong, Boyuan Jiang, Donghao Luo, Wenqing Chu, Xiaoming Huang, Ying
  Tai, Chengjie Wang, and Jie Yang.
\newblock Ifrnet: Intermediate feature refine network for efficient frame
  interpolation.
\newblock In {\em Proceedings of the IEEE/CVF Conference on Computer Vision and
  Pattern Recognition}, pages 1969--1978, 2022.

\bibitem{lee2020adacof}
Hyeongmin Lee, Taeoh Kim, Tae-young Chung, Daehyun Pak, Yuseok Ban, and
  Sangyoun Lee.
\newblock Adacof: Adaptive collaboration of flows for video frame
  interpolation.
\newblock In {\em Proceedings of the IEEE/CVF Conference on Computer Vision and
  Pattern Recognition}, pages 5316--5325, 2020.

\bibitem{deepvoxelflow}
Ziwei Liu, Raymond~A. Yeh, Xiaoou Tang, Yiming Liu, and Aseem Agarwala.
\newblock Video frame synthesis using deep voxel flow.
\newblock In {\em The IEEE International Conference on Computer Vision (ICCV)},
  Oct 2017.

\bibitem{long2016learning}
Gucan Long, Laurent Kneip, Jose~M Alvarez, Hongdong Li, Xiaohu Zhang, and
  Qifeng Yu.
\newblock Learning image matching by simply watching video.
\newblock In {\em European Conference on Computer Vision}, pages 434--450.
  Springer, 2016.

\bibitem{VFIformer}
Liying Lu, Ruizheng Wu, Huaijia Lin, Jiangbo Lu, and Jiaya Jia.
\newblock Video frame interpolation with transformer.
\newblock In {\em Proceedings of the IEEE/CVF Conference on Computer Vision and
  Pattern Recognition}, pages 3532--3542, 2022.

\bibitem{Niklaus2018CVPR}
Simon Niklaus and Feng Liu.
\newblock Context-aware synthesis for video frame interpolation.
\newblock In {\em The IEEE Conference on Computer Vision and Pattern
  Recognition (CVPR)}, June 2018.

\bibitem{niklaus2020softmax}
Simon Niklaus and Feng Liu.
\newblock Softmax splatting for video frame interpolation.
\newblock In {\em Proceedings of the IEEE/CVF Conference on Computer Vision and
  Pattern Recognition}, pages 5437--5446, 2020.

\bibitem{adaconv}
Simon Niklaus, Long Mai, and Feng Liu.
\newblock Video frame interpolation via adaptive convolution.
\newblock In {\em The IEEE Conference on Computer Vision and Pattern
  Recognition (CVPR)}, July 2017.

\bibitem{sepconv}
Simon Niklaus, Long Mai, and Feng Liu.
\newblock Video frame interpolation via adaptive separable convolution.
\newblock In {\em The IEEE International Conference on Computer Vision (ICCV)},
  Oct 2017.

\bibitem{paikin2021efi}
Genady Paikin, Yotam Ater, Roy Shaul, and Evgeny Soloveichik.
\newblock Efi-net: Video frame interpolation from fusion of events and frames.
\newblock In {\em Proceedings of the IEEE/CVF Conference on Computer Vision and
  Pattern Recognition}, pages 1291--1301, 2021.

\bibitem{park2020bmbc}
Junheum Park, Keunsoo Ko, Chul Lee, and Chang-Su Kim.
\newblock Bmbc: Bilateral motion estimation with bilateral cost volume for
  video interpolation.
\newblock In {\em Computer Vision--ECCV 2020: 16th European Conference,
  Glasgow, UK, August 23--28, 2020, Proceedings, Part XIV 16}, pages 109--125.
  Springer, 2020.

\bibitem{ABME}
Junheum Park, Chul Lee, and Chang-Su Kim.
\newblock Asymmetric bilateral motion estimation for video frame interpolation.
\newblock In {\em Proceedings of the IEEE/CVF International Conference on
  Computer Vision}, pages 14539--14548, 2021.

\bibitem{PontTusetarXiv2017}
Jordi Pont-Tuset, Federico Perazzi, Sergi Caelles, Pablo Arbel\'aez, Alexander
  Sorkine-Hornung, and Luc {Van Gool}.
\newblock The 2017 davis challenge on video object segmentation.
\newblock {\em arXiv:1704.00675}, 2017.

\bibitem{ronneberger2015u}
Olaf Ronneberger, Philipp Fischer, and Thomas Brox.
\newblock U-net: Convolutional networks for biomedical image segmentation.
\newblock In {\em International Conference on Medical image computing and
  computer-assisted intervention}, pages 234--241. Springer, 2015.

\bibitem{VFIT}
Zhihao Shi, Xiangyu Xu, Xiaohong Liu, Jun Chen, and Ming-Hsuan Yang.
\newblock Video frame interpolation transformer.
\newblock In {\em Proceedings of the IEEE/CVF Conference on Computer Vision and
  Pattern Recognition}, pages 17482--17491, 2022.

\bibitem{siyao2021deep}
Li Siyao, Shiyu Zhao, Weijiang Yu, Wenxiu Sun, Dimitris Metaxas, Chen~Change
  Loy, and Ziwei Liu.
\newblock Deep animation video interpolation in the wild.
\newblock In {\em Proceedings of the IEEE/CVF Conference on Computer Vision and
  Pattern Recognition}, pages 6587--6595, 2021.

\bibitem{soomro2012ucf101}
Khurram Soomro, Amir~Roshan Zamir, and Mubarak Shah.
\newblock Ucf101: A dataset of 101 human actions classes from videos in the
  wild.
\newblock {\em arXiv preprint arXiv:1212.0402}, 2012.

\bibitem{sun2018pwc}
Deqing Sun, Xiaodong Yang, Ming-Yu Liu, and Jan Kautz.
\newblock Pwc-net: Cnns for optical flow using pyramid, warping, and cost
  volume.
\newblock In {\em The IEEE Conference on Computer Vision and Pattern
  Recognition (CVPR)}, June 2018.

\bibitem{teed2020raft}
Zachary Teed and Jia Deng.
\newblock Raft: Recurrent all-pairs field transforms for optical flow.
\newblock In {\em European conference on computer vision}, pages 402--419.
  Springer, 2020.

\bibitem{wang2004image}
Zhou Wang, Alan~C Bovik, Hamid~R Sheikh, Eero~P Simoncelli, et~al.
\newblock Image quality assessment: from error visibility to structural
  similarity.
\newblock {\em IEEE transactions on image processing}, 13(4):600--612, 2004.

\bibitem{wu2018video}
Chao-Yuan Wu, Nayan Singhal, and Philipp Krahenbuhl.
\newblock Video compression through image interpolation.
\newblock In {\em Proceedings of the European Conference on Computer Vision
  (ECCV)}, pages 416--431, 2018.

\bibitem{xu2019quadratic}
Xiangyu Xu, Li Siyao, Wenxiu Sun, Qian Yin, and Ming-Hsuan Yang.
\newblock Quadratic video interpolation.
\newblock {\em Advances in Neural Information Processing Systems},
  32:1647--1656, 2019.

\bibitem{xue2019video}
Tianfan Xue, Baian Chen, Jiajun Wu, Donglai Wei, and William~T Freeman.
\newblock Video enhancement with task-oriented flow.
\newblock {\em International Journal of Computer Vision}, 127(8):1106--1125,
  2019.

\bibitem{Yu2021ICCV}
Zhiyang Yu, Yu Zhang, Deyuan Liu, Dongqing Zou, Xijun Chen, Yebin Liu, and
  Jimmy~S. Ren.
\newblock Training weakly supervised video frame interpolation with events.
\newblock In {\em Proceedings of the IEEE/CVF International Conference on
  Computer Vision (ICCV)}, pages 14589--14598, October 2021.

\bibitem{zhang2018unreasonable}
Richard Zhang, Phillip Isola, Alexei~A Efros, Eli Shechtman, and Oliver Wang.
\newblock The unreasonable effectiveness of deep features as a perceptual
  metric.
\newblock In {\em Proceedings of the IEEE conference on computer vision and
  pattern recognition}, pages 586--595, 2018.

\bibitem{zhou2016view}
Tinghui Zhou, Shubham Tulsiani, Weilun Sun, Jitendra Malik, and Alexei~A Efros.
\newblock View synthesis by appearance flow.
\newblock In {\em European conference on computer vision}, pages 286--301.
  Springer, 2016.

\end{thebibliography}
}

\clearpage
\clearpage

\appendix
\label{sec:appendix}
\section*{\Large{Appendices}}
\section{Network Details}
\label{sec:sub_network}
We employ three Video Frame Interpolation~(VFI) models (AdaCoF~\cite{lee2020adacof}, CAIN~\cite{choi2020channel}, VFIT~\cite{VFIT}) as baselines to demonstrate the effectiveness of our methods. Since the network architectures are different, we apply our $D$-map estimators to them in suitable ways respectively.
The module of estimating $D$-map is basically taking general structure of decoder.

AdaCoF uses a U-Net architecture and obtains parameters after passing to the network for interpolation. Therefore, we take a specific layer for  the discontinuity map~($D$-map) from the decoded feature~(the output of the 18th layer of its paper). The size of the feature map~($\mathcal{F}$) is $1/2$ the original size, and its channel size is 128.

VFIT is similar to AdaCoF, however, they use a hierarchical structure for the decoder. To estimate $D$-map without affecting the performance of interpolation, we select a specific layer from the highest-level feature~(the output of the 17th layer in the paper). The size of the feature map~($\mathcal{F}$) is $1/16$ the size of the original, and its channel size is 2048.
 
CAIN takes simple architecture for interpolation with pixel-shuffle and attention. We obtain feature map from the output of the encoder~(the output of the first convolution layer in the paper). The size of the feature map~($\mathcal{F}$) is $1/8$ the original size, and its channel size is 384. Consequently, we successfully employ various baseline algorithms for expanding motions by applying our methods in different ways.

\section{Network Complexity}
Table~\ref{tbl:sub_Dmap} shows the additional complexity when our methods are applied.
As we mentioned in Section~\ref{sec:sub_network}, the numbers of specific layer for the discontinuity map are different respectively in each network, so the complexity for baseline networks are slightly different.
The complexity can be decreased by adjusting the position where we extract the features for discontinuity map. 

\vspace{-0.3cm}
\begin{table}[h]
\setlength{\belowcaptionskip}{-20pt}
	\begin{center}
	\resizebox{0.99\columnwidth}{!}{
			\begin{tabular}{lccc}
				\toprule
				& AdaCoF & CAIN & VFIT   \\
				\midrule
                GFlops & 49.7/ 53.4 (+6.9\%) & 88.5 / 114.4 (+22.6\%) & 456 / 521(+12.4\%)\\
				\bottomrule
		\end{tabular}}
		\vspace{-0.3cm}
		\caption{The model complexity analysis}
		\label{tbl:sub_Dmap}
	\end{center}
\end{table}

\section{Graphical Discontinuous Motion dataset}
\label{sec:sub_GDM}
We construct a new test set called Graphic Discontinuous Motion~(GDM) dataset. The GDM dataset consists of high-resolution videos obtained from three types of games. It has 30 sequences and each sequence has $1920\times 1080$ resolution. Its videos contain not only continuous motions but also discontinuous motions.

\begin{figure}[t]
	\setlength{\belowcaptionskip}{-15pt}
	\begin{center}
		\includegraphics[width=0.99\linewidth]{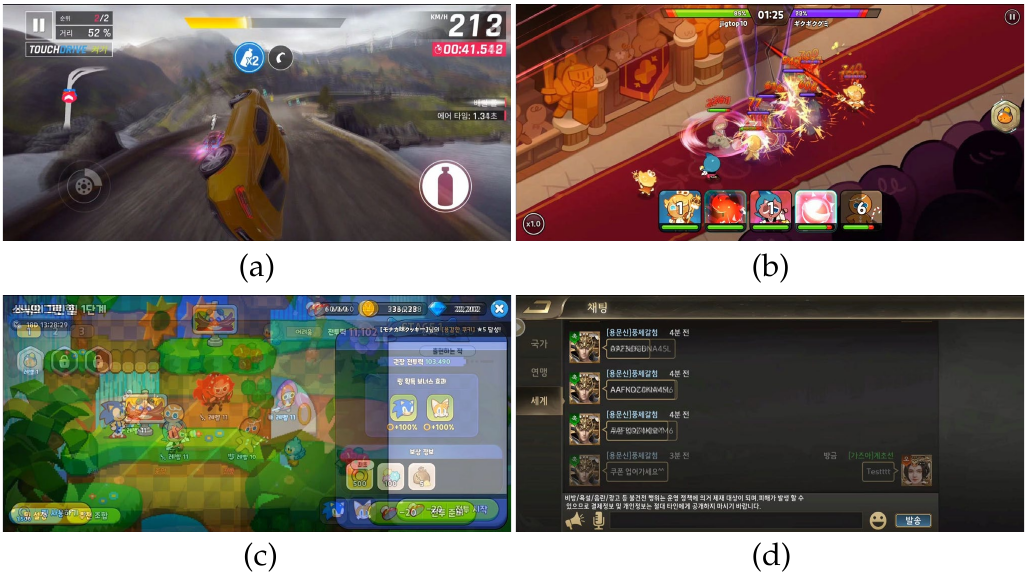}
	\end{center}
	\vspace{-13pt}
	\caption{Overlaid examples of GDM dataset}
	\label{fig:sub_GDM}
\end{figure}

Figure~\ref{fig:sub_GDM} shows the part of the GDM dataset that contains various types of discontinuous motions. In Figure~\ref{fig:sub_GDM}, (a) and (b) show the example of the user interface and the numbers. Case (c) represents when the scene suddenly switches. (d) is the chatting example when we usually do streaming or gaming. As shown in Figure~\ref{fig:sub_GDM}, we find that the GDM dataset can be used for evaluating the performance of interpolation without bias.

\begin{figure*}[t]
	\setlength{\belowcaptionskip}{-15pt}
	\begin{center}
		\includegraphics[width=0.99\linewidth]{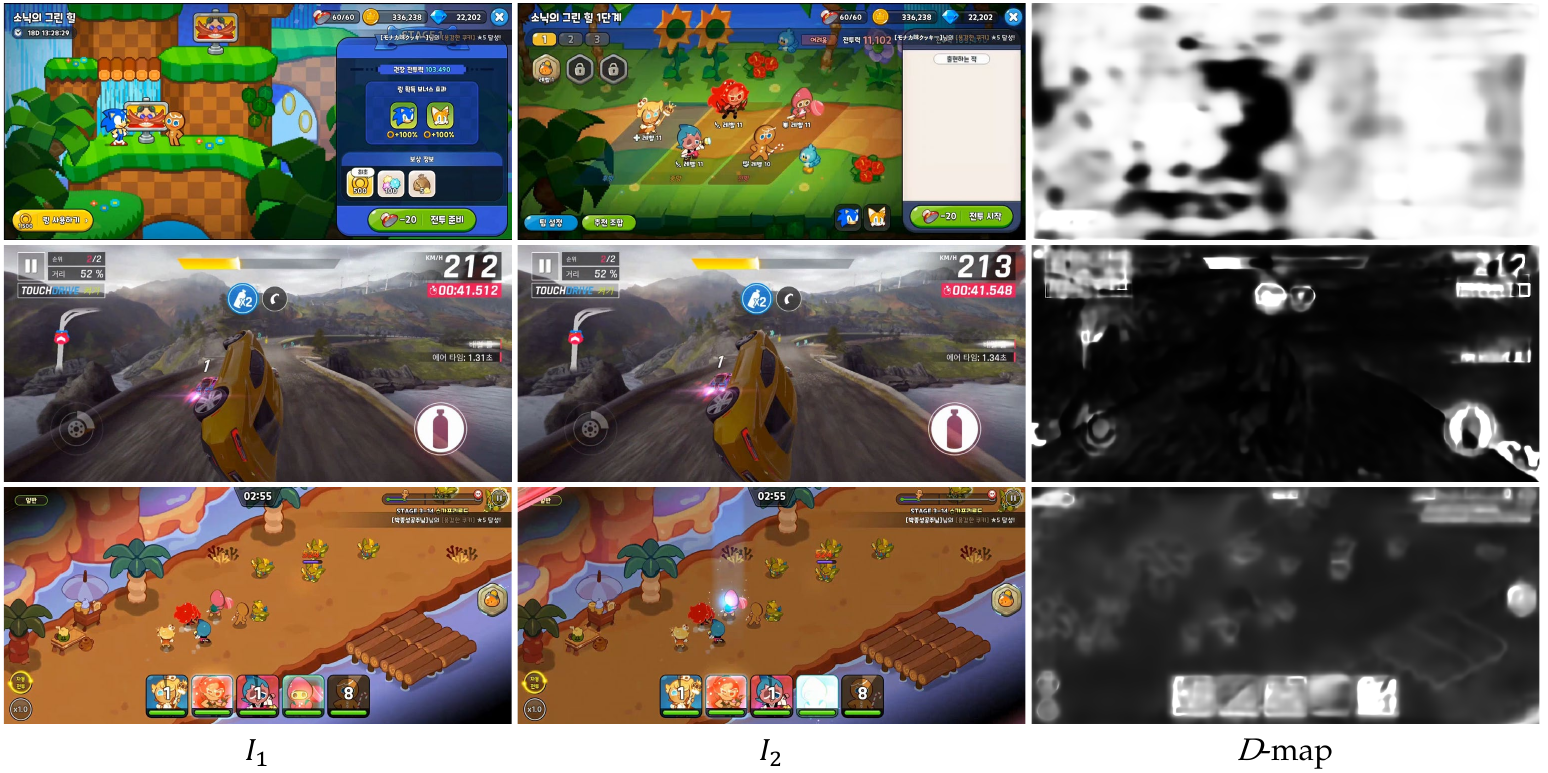}
	\end{center}
	\vspace{-13pt}
	\caption{The examples of $D$-map in various videos.}
	\label{fig:sub_dmap}
\end{figure*}

\section{Figure-Text Mixing Details}
\label{sec:sub_FTM}
As general data augmentation, we randomly crop the 256 × 256 patches and flip them horizontally, vertically, and temporally for training. Then, we add Figure-Text Mixing~(FTM) augmentation which consists of Figure Mixing and Text Mixing. For each mixing technique, we present implementation details in Section~\ref{sec:sub_FM} and \ref{sec:sub_TM}.

\subsection{Figure Mixing}
\label{sec:sub_FM}
Figure Mixing~(FM) contains two types of figure, square and circle. Each type can be added respectively. The probability of decision for adding FM is 0.5. Details of FM are shown below.

\begin{itemize}
\item \textbf{type:} square / circle
\item \textbf{size:} $10 \leq$ \textit{height, weight} $\leq 41$
\item \textbf{color:} random RGB colors
\item \textbf{thickness:} $1 \leq$ \textit{thickness} $\leq 4$
\item \textbf{position:} same position for entire videos
\end{itemize}

\subsection{Text Mixing}
\label{sec:sub_TM}
Text Mixing~(TM) presents various types of augmentation. Unlike FM, TM consists of four cases: 1) the position of the text is fixed in the entire video, 2) the text does not exist in the previous frame and appears in the future frame, 3) the existing text suddenly disappears, 4) the text moves up and down by its vertical size. The details of TM are shown below.
\begin{itemize}
\item \textbf{type:} random text
\item \textbf{text length:}  $5 \leq$ \textit{length} $\leq 30$
\item \textbf{font size:} $10 \leq$ \textit{size} $\leq 40$
\item \textbf{font type:} windows basic fonts
\item \textbf{color:} random RGB colors
\item \textbf{position:} same position for input frames
\end{itemize}

\begin{figure*}[t]
	\setlength{\belowcaptionskip}{-15pt}
	\begin{center}
		\includegraphics[width=0.99\linewidth]{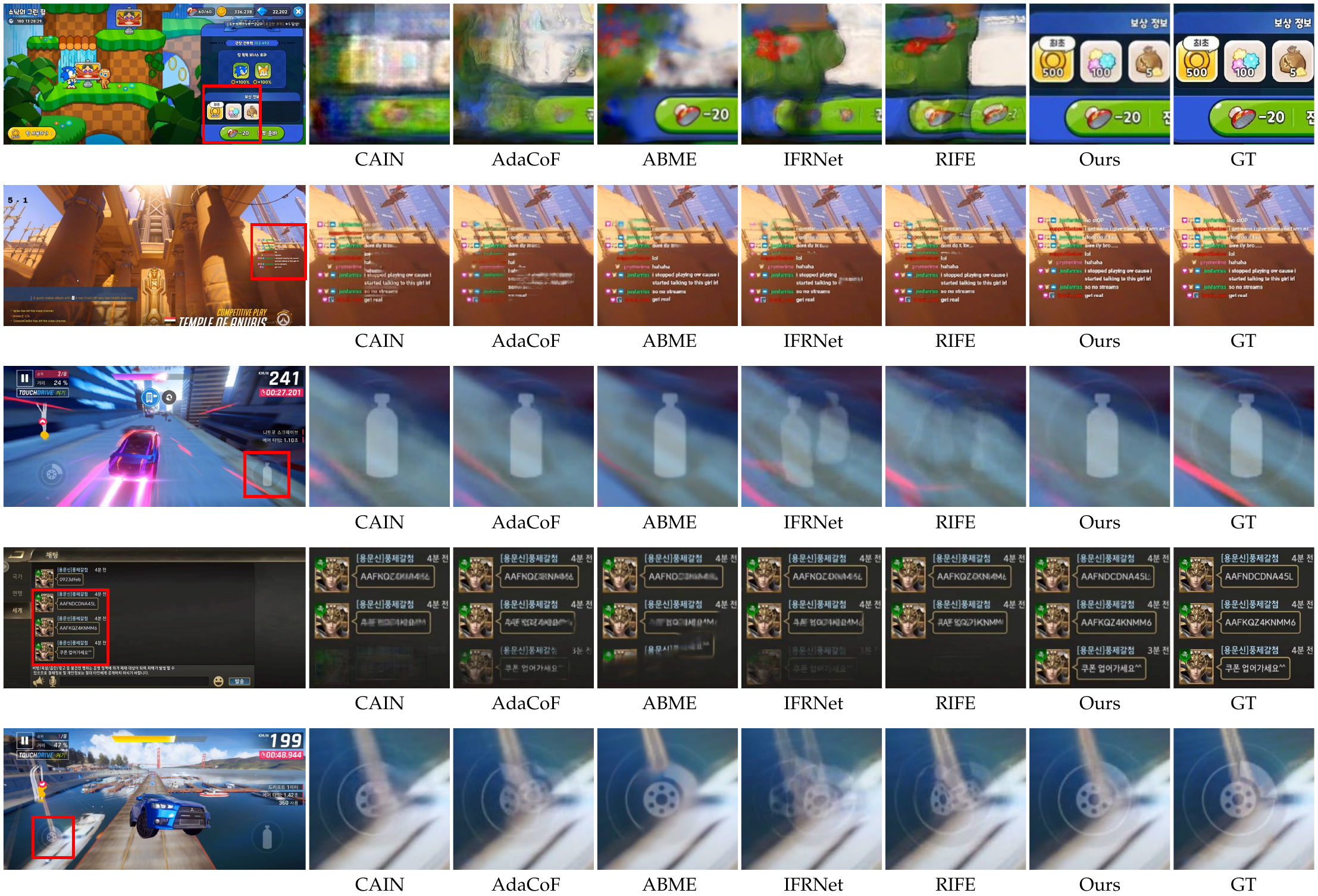}
	\end{center}
	\vspace{-13pt}
	\caption{Visual comparison of with discontinuous motion.}
	\label{fig:sub_discontinuous}
\end{figure*}

\section{Additional Discontinuity Map Visualizations}
\label{sec:sub_dmap}
The discontinuity map~($D$-map), which highlights discontinuously moving objects in the input image, is an important idea to deal with discontinuous motions. Figure~\ref{fig:sub_dmap} shows the visualizations of $D$-maps, where $I_1$ and $I_2$ are the previous and next adjacent frames. The first row of Figure~\ref{fig:sub_dmap} shows an example of an  immediately changing scene. In this case, the $D$-map highlights almost  the entire area that needs to be copied from the previous frame. The second and third rows of Figure~\ref{fig:sub_dmap} demonstrate the effectiveness of the $D$-map. $D$-map successfully separates the discontinuous regions from the entire frames containing both continuous and discontinuous motion. Therefore, these results prove that $D$-map plays a role to suit its purpose, which estimates the regions that should copy and paste from the previous frame. It is especially impressive that the highlighted objects are not in the training dataset and also cannot be learned even with FTM augmentation.

\section{Additional Qualitative Results}
\label{sec:sub_Qualitative}
We show additional results for discontinuous motion in Figure~\ref{fig:sub_discontinuous}. The 1st row in Figure~\ref{fig:sub_discontinuous} is an example of immediate scene transition. The previous algorithms force interpolation of the two unrelated input frames, resulting in damaged frames. However, our method shows a clear result, almost the same as the ground truth image. The 2nd and 4th rows represent the typical discontinuous motion, the scenario of a chatting window. The text moves up discontinuously. The previous methods produce overlapped or distorted frames. However, our method catches the discontinuous motion and shows clear results.
3rd and 5th rows are examples of static user interfaces that can be frequently found in many games. The previous results usually fail, especially when there are some large motions near them. However, our method robustly maintains the structure of the user interfaces better than previous algorithms.

\end{document}